\documentclass[10pt,twocolumn,letterpaper]{article}

\pdfoutput=1

\usepackage{iccv}
\usepackage{times}
\usepackage{epsfig}
\usepackage{graphicx}
\usepackage{amsmath}
\usepackage{amssymb}

\usepackage[T1]{fontenc}

\usepackage[breaklinks=true,bookmarks=false]{hyperref}

\iccvfinalcopy 

\def\iccvPaperID{2600} 

\ificcvfinal\fi

\begin{document}


\title{CSG-Stump: A Learning Friendly CSG-Like Representation \\ for Interpretable Shape Parsing}

\author{Daxuan Ren$^{1,2}$ \quad
Jianmin Zheng$^{1, \dagger}$ \quad
Jianfei Cai$^{3}$ \quad
Jiatong Li$^{1,2}$ \quad
Haiyong Jiang$^{5}$ \quad
\\Zhongang Cai$^{2,4}$ \quad
Junzhe Zhang$^{2,4}$ \quad
Liang Pan$^{4}$ \quad
Mingyuan Zhang$^{2,4}$ \quad
Haiyu Zhao$^{2}$ \quad
Shuai Yi$^{2}$ \quad
\\
$^{1}$Nanyang Technological University (NTU)  \hspace{8pt} $^{2}$SenseTime Research \hspace{8pt} $^{3}$Monash University\\
$^{4}$ S-Lab, NTU \hspace{8pt} $^{5}$ University of Chinese Academy of Sciences\\

{\tt\small {\{daxuan001, asjmzheng, E180176, haiyong.jiang, junzhe001, liang.pan\}@ntu.edu.sg}}\\
{\tt\small {\{zhangmingyuan, caizhongang, zhaohaiyu, yishuai\}@sensetime.com}}  \tt\small   {jianfei.cai@monash.edu}\\
}
\maketitle
\ificcvfinal\thispagestyle{empty}\fi

\begin{abstract}
    Generating an interpretable and compact representation of 3D shapes from point clouds is an important and challenging problem. This paper presents CSG-Stump Net, an unsupervised end-to-end network for learning shapes from point clouds and discovering the underlying constituent modeling primitives and operations as well. At the core is a three-level structure called {\em CSG-Stump}, consisting of a complement layer at the bottom, an intersection layer in the middle, and a union layer at the top. 
    CSG-Stump is proven to be equivalent to CSG in terms of representation, therefore inheriting the interpretable, compact and editable nature of CSG while freeing from CSG's complex tree structures. Particularly, the CSG-Stump has a simple and regular structure, allowing neural networks to give outputs of a constant dimensionality, which makes itself deep-learning friendly.
  Due to these characteristics of CSG-Stump, CSG-Stump Net achieves superior results compared to previous CSG-based methods and generates much more appealing shapes, as confirmed by extensive experiments.
  
  \let\thefootnote\relax\footnotetext{Project Page: \href{https://kimren227.github.io/projects/CSGStump/}{https://kimren227.github.io/projects/CSGStump/}}
\let\thefootnote\relax\footnotetext{$\dagger$ Indicates corresponding author}
 \let\thefootnote\relax\footnotetext{The work is partially supported by a joint WASP/NTU project (04INS000440C130).}
\end{abstract}

\section{Introduction}

\begin{figure}[th]
  \centering
  \includegraphics[width=0.4\textwidth]{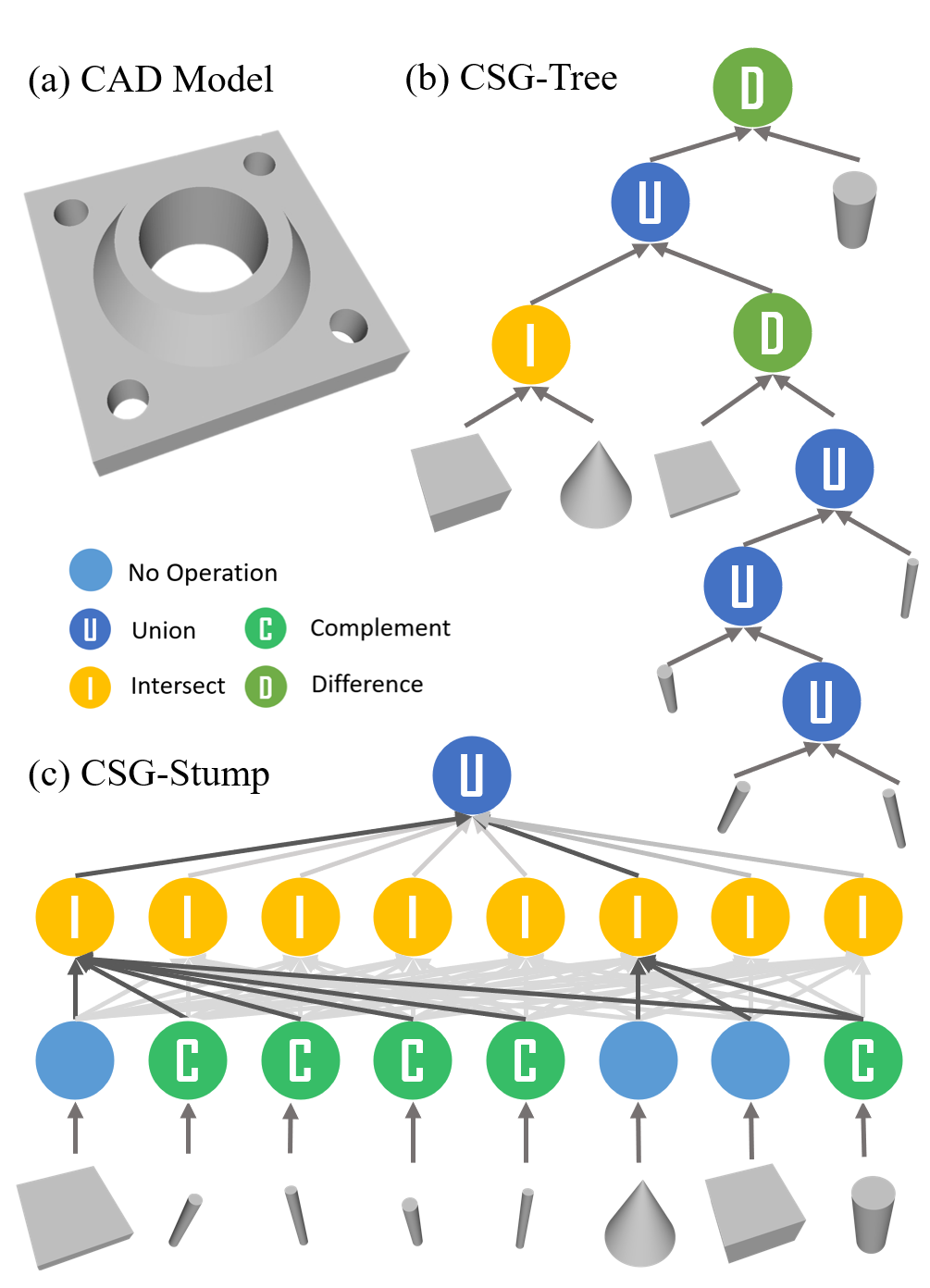}
  \caption{A CAD model (a) can be represented as either a CSG representation (b) or a CSG-Stump representation (c). CSG-Stump is equivalent to CSG but frees from CSG's irregular tree structure. Thus CSG-Stump is more friendly to optimization formulation and network designs. Here nodes ``I'', ``U'', ``D'', and ``C''  denote intersection, union, difference, and (shape) complement, respectively.}
  \label{fig:teaser}
  \vspace{-0.1in}
\end{figure}
Shape is a geometric form, which helps us understand objects, surrounding environments and even the world. Therefore shape modeling and understanding has always been a research topic in computer vision and graphics. Various representations have been developed for 3D shapes. Examples are point clouds~\cite{qi2017pointnet, qi2017pointnet++, wang2019dynamic, thomas2019kpconv}, 3D voxels~\cite{wang2018adaptive}, implicit fields~\cite{mescheder2019occupancy, park2019deepsdf, chen2019learning, hao2020dualsdf, chen2019bae, saito2019pifu, mildenhall2020nerf},  meshes~\cite{guo20153d, bronstein2017geometric, wang2018pixel2mesh, pontes2018image2mesh, wen2019pixel2mesh++}, and parametric representations~\cite{sharma2018csgnet,kania2020ucsg}.  
With the advance in 3D acquisition technologies, point clouds are easily generated, but they are a set of unstructured points and lack explicit high-level structure and semantic information. There is a great demand for converting point clouds to high-level shape representations that help recognize and understand the shapes, supporting the designer to re-create new products and facilitate various applications such as building the digital twins of products and systems~\cite{LU20171}. Particularly, reverse engineering (RE) technologies, especially reconstructing implicit or parametric (CAD) models from point clouds, have been extensively studied in engineering. However, most prior art involves  a tedious and time-consuming process and has difficulty in fully addressing the requirements of the industry, which actually indicates a need of a paradigm shift.  

In recent years, deep learning has achieved substantial success in areas such as computer vision and natural language processing and shows great potential in solving complex problems that are difficult to be solved with traditional techniques. The exploration of deep learning techniques for high-level shape reconstruction from point clouds also gains much popularity. In particular, a few works exploit neural network techniques for parsing point cloud models into their Constructive Solid Geometry (CSG) tree \cite{laidlaw1986constructive}, which is a widely used 3D representation and modeling processing in the CAD industry. CSG models a shape by iteratively performing Boolean operations on simple parametric primitives, usually followed by a binary tree (see Fig.~\ref{fig:teaser}). Thus CSG is an ideal model for providing compact representation, high interpretability, and editability. 
However, the binary CSG-Tree structure introduces two challenges: 
1) it is difficult to define a CSG-Tree with a fixed dimension formulation; 2) the iterative nature of CSG-Tree construction cannot be formulated as matrix operations and a long sequence optimization suffers varnishing gradients.

CSG-Net~\cite{sharma2018csgnet} pioneers deep learning based CSG parsing by employing an RNN for the tree structure prediction. However, CSG-Net requires expensive annotations with expert knowledge, which is difficult to scale. BSP-Net~\cite{chen2020bsp} and CVX-Net~\cite{deng2020cvxnet} propose to leverage a set of parametric hyperplanes to represent a shape, but abundant hyperplanes are needed to approximate curved surfaces. Overall, these methods are still not efficient, interpretable, or easy-editable. More importantly, these methods assume a frozen combination among predicted hyperplanes during inference, which effectively collapses into a fixed order of operations, limiting its theoretical representability.  
UCSG-Net\cite{kania2020ucsg} proposes the CSG-Layer to generate highly interpretable shapes by a multi-layer CSG-Tree iteratively, but only a few layers can be supported (five layers in UCSG-Net) because of the optimization difficulty, which greatly restricts the diversity and representation capability.

In this paper, we propose {\em CSG-Stump}, a novel and systematic reformulation of CSG-Tree. CSG-Stump has a fixed tree structure of only three layers (hence the name \textit{stump}). We prove that CSG-Stump is equivalent to typical CSG-Tree in terms of representation, i.e., we can represent any complex CSG shape by our three-layer CSG-Stump (see Fig.\ref{fig:teaser}). Therefore, CSG-Stump inherits the ideal characteristics of CSG-Tree, allowing highly compact, interpretable and editable shape representation while freeing from the limitations of a tree structure. Moreover, CSG-Stump gives rise to two additional advantages: 1) High representation capability. The maximum representation capability can be realistically achieved with CSG-Stump, as opposed to a conventional CSG-Tree that needs many layers for complex shapes. 2) Deep learning-friendly. The consistent structure of CSG-Stump allows neural networks to give fixed dimension output, making network design much easier. 

We also propose two methods to automatically construct CSG-Stump from unstructured raw inputs, e.g., point clouds. The first approach is to detect basic primitives using off-the-shelf methods, e.g. RANSAC \cite{schnabel2007efficient}, and then convert the problem to a Binary Programming problem to estimate the primitive constructive relations.
To overcome the issues such as precision requirements on the inputs, manual parameter tuning and scalability due to the combinational nature of the problem,  we in the second approach design a simple end-to-end network for joint primitive detection and CSG-Stump estimation (see Sec.~\ref{sec:CSGStumpNet}). This data-driven approach is more efficient. Moreover, it can learn useful priors for primitive detection and assembly from large scale data. Notably, this network is trained in an unsupervised manner, i.e., without the need of expensive annotations of CSG parsing trees from trained professionals.
Experimental results show that our CSG-Stump exhibits remarkable representation capability while preserving the interpretable, compact and editable nature of CSG representation.

In summary, the paper has the following contributions:
\begin{itemize}
    \vspace{-2mm}
    \setlength\itemsep{-1mm}
    \item We propose CSG-Stump, a three-layer reformulation of the classic CSG-Tree for a better interpretable, trainable and learning friendly representation, and provide theoretical proof of the equivalence between CSG-Stump and CSG-Tree.
    \item We demonstrate that CSG-Stump is highly compatible with deep learning. With its help, even a simple unsupervised end-to-end network can perform dynamic shape abstraction. 
    \item Extensive experiments are conducted to show that CSG-Stump achieves state-of-the-art results both quantitatively and qualitatively while allowing further edits and manipulation.
\end{itemize}

\section{Related Work}

This section briefly reviews 3D shape representation, neural point cloud learning and high-level shape parsing and reconstruction, which are relevant to our work.

\noindent \textbf{3D Shape Representation.}  
In 3D computer vision, diverse representations are designed and proposed for different applications, each containing its own advantages and drawbacks. Point cloud is the widely adopted raw input format for 3D data due to its flexibility in representing shape details and wide usages in data collection~\cite{qi2017pointnet,qi2017pointnet++,wang2019dynamic}, but its unstructured nature makes it hard to edit. Mesh is simple to use and render, but its variant topology requires additional processes for learning~\cite{guo20153d, bronstein2017geometric}. Volumetric representation extends 2D grid representation of images to 3D voxels, making it easy to incorporate innovative network designs in 2D, but the memory and computation-hungry nature limit its resolutions, leading to the lack of geometry details~\cite{wu20153d, riegler2017octnet, wang2018adaptive, wang2017cnn, choy20163d}. Implicit representation frees from topology and resolution issues, but heavy computation is required for tessellation and mesh generation~\cite{mescheder2019occupancy, park2019deepsdf, hao2020dualsdf, chen2019bae, chen2019learning, saito2019pifu, mildenhall2020nerf}.
Moreover, these representations do not account for the structural and semantic organization of 3D shapes.

\noindent \textbf{Point Cloud Learning.}
3D raw inputs are usually in the form of point clouds. Qi et al.~\cite{qi2017pointnet,qi2017pointnet++} pioneered 3D deep learning on point clouds by introducing permutation-invariant feature learning and multi-scale feature aggregation. Wang et al.~\cite{wang2019dynamic} explored neighborhood information via a dynamically constructed graph and edge convolution. More recently, Thomas et al.  \cite{thomas2019kpconv} proposed Kernel Point as a new convolution operator and achieved a state-of-the-art result on common benchmarks. In this paper, we focus on structural shape fitting instead of point cloud feature extraction. In particular, for simplicity, we use DGCNN~\cite{wang2019dynamic} as our backbone, and minimal effort is required to swap to other point cloud encoders.

\noindent \textbf{High-Level Shape Parsing and Reconstruction.} Recently, there has been an increasing interest in parsing the shape to its high-level representations. 
For example, generating parametric shapes using data-driven methods has gained popularity. 
Tulsiani et al.~\cite{tulsiani2017learning} pioneered deep learning based parametric shape representation by abstracting a shape as a union of boxes. 
Paschalidou et al.~\cite{paschalidou2019superquadrics} employed superquadrics instead of boxes as basic primitives to achieve better approximation. Both methods only support the union operation, which limits the representation capability.
In contrast to solid primitives, Li et al. proposed SPFN~\cite{li2019supervised}, a supervised method for parametric surface prediction. SPFN does not consider the mutual relations among the predicted primitives, which leads to improperly generated shapes.
BSP-Net~\cite{chen2020bsp} and CVX-Net~\cite{deng2020cvxnet} achieved remarkable results by exploring half-space partition. These two methods require a large number of planes to approximate non-planar surfaces. Hence, though parametric, they are still less interpretable.

CSG~\cite{laidlaw1986constructive} is a modeling procedure and a representation for 3D Shapes as well. CSG is widely used in industrial software like SolidWorks \cite{solidworks2005solidworks} and OpenSCAD \cite{kintel2014openscad} due to its intuitive and powerful concept. In \cite{goldfeather1989near}, a normalized CSG representation was proposed for fast rendering by rearranging CSG operations into union of intersections. However, the normalized CSG is still a tree structure with varying depths, which makes it difficult to be directly inferred by a neural network.
Recently, a few methods have been proposed to tackle this problem. Sharma et al.~\cite{sharma2018csgnet} used RNN to generate a sequence of primitives and operations in a supervised manner and then parsed the sequence as a CSG-Tree. Annotating parsing trees for a large corpus of 3D shapes however requires professional knowledge and tedious annotation processes. 
UCSG-Net~\cite{kania2020ucsg} took an unsupervised approach but required iterative operand selections for each tree branch. This iterative process makes it hard to extend to a very deep structure (only $5$ levels in the paper) due to gradient vanishing. 

Our proposed CSG-Stump \emph{squashes} a CSG-Tree of arbitrary depth into a fixed three-layer representation and uses connection matrices to represent variations in different CSG relations. This regular structure alleviates the problem of handling tree structures and makes it much easier to be incorporated in a network.

\section{CSG-Stump}\label{sec:CSGStump}
This section presents CSG-Stump, a three-layer tree representation for 3D shapes. At the top is a {\em union} layer with only one node. In the middle is an {\em intersection} layer, and at the bottom is a {\em complement} layer (see Fig.~\ref{fig:teaser}). CSG-Stump also contains a set of primitive objects.
The nodes at the complement layer correspond to the primitives one-to-one.  Nodes at different layers contain some information for operations, as indicated by their names. Specifically, the nodes at the complement layer store whether the complement operation is performed on their corresponding primitives. The nodes at the intersection layer record which shapes generated in the bottom layer are selected for the intersection operation. The node at the top layer records which shapes generated in the intersection layer are selected for the union operation.

To facilitate discussion and analysis, we also introduce two special shapes as primitives: the whole space and the empty set denoted by $\mathcal{U}$ and $\emptyset$, respectively. The complement of a shape is implemented by the difference of the shape with the whole space. Intersecting an object with $\mathcal{U}$ gives the object itself. Similarly, a union of an object and $\emptyset$ also returns the object itself. 

For each layer in the CSG-Stump, we introduce a connection matrix to encode the information for its nodes. Specifically, we define a $1\times K$ matrix $W_C\in\{0,1\}^K$ for the complement layer, where $K$ is the number of the primitives; a $K\times C$ matrix $W_I\in \{0,1\}^{K\times C}$ for the intersection layer, where $C (\le K)$ is the number of nodes in the intersection layer; and a $C\times 1$ matrix
$W_U \in \{0,1\}^{C}$ for the union layer. 
Each entry in these matrices takes a value of either 1 or 0. In particular,  $W_C[1,i]=0$ or $1$ encodes whether the shape of primitive $i$ or its complement is used for node $i$ of the complement layer. If $W_I[j,i] = 1$, the shape from node $j$ in the complement layer is selected for the intersection in node $i$ of the intersection layer, and similarly, $W_U[j,1]$ implies that the shape from node $j$ in the intersection layer is selected for the union operation at the top layer. In this way, CSG-Stump represents a shape by a set of primitive shapes and three connection matrices.   

Similar to the CSG-Tree, CSG-Stump is a hierarchical representation with nodes storing operation information, but with fixed three layers. The CSG-Tree representation is usually organized as a binary tree with many layers, which makes the prediction of primitives and Boolean operations a tedious and challenging iterative process~\cite{kania2020ucsg,sharma2018csgnet}. Particularly, working with a long sequence not only causes problems in gradient feedback but also is sensitive to the order of primitives and Boolean operations. In contrast, CSG-Stump has a fixed type of Boolean operations at each layer and only requires determining three binary connection matrices, which makes CSG-Stump learning friendly.  

\subsection{Function Representation} \label{CSG_Operations}

To facilitate shape analysis and problem formulation, we describe the nodes of CSG-Stump by mathematics functions. First, we define a shape $O$ by an occupancy function $O(x):\mathbb{R}^3 \rightarrow \{0,1\}$ as follows:

\begin{equation}
O(x) = \left\{
\begin{array}{ll}
1, & x \;\;\text{is within the shape}\\
0, & \text{otherwise}
\end{array}
\right.
\end{equation}
which encodes the occupancy of the shape in 3D space. Here we use $O$ for both the shape and its occupancy function, and we adopt the same convention in the rest of the paper where there is no ambiguity. Thus $\mathcal{U}(x) \equiv 1$ and $\emptyset(x) \equiv 0$.
The Boolean operations can then be formulated by simple mathematics functions. In particular, the complement of primitive object $O_i$ can be defined by function $O_i^c(x)= 1-O_i(x)$, the intersection of $k$ objects, $\bigcap_i^k O_i$, by $\min_{i = 1 \dots k}(O_i(x))$, the union  of $k$ objects, $\bigcup_i^k O_i$, by $ \max_{i = 1 \dots k}(O_i(x))$, and the difference $O_i-O_j$ of two objects $O_i$ and $O_j$ by $\min\left(O_i(x), 1-O_j(x)\right)$.

With the binary connection matrices $W_C$, $W_I$ and $W_U$,
each node in the CSG-Stump structure can be defined by a certain function. 
\begin{itemize}
\item For each node $i = 1,\cdots, K$ in the first layer, its shape $F_i$ can be defined by function $F_i(x)$:
\begin{equation}\label{eq:Fi}
    F_i(x) = W_C[1,i]\times (1-O_i(x)) + (1-W_C[1,i])O_i(x).
\end{equation}
\item For each node $i = 1,\cdots, C$ in the second layer, its shape $S_i$ is an intersection of nodes from the first layer, $ {\bigcap_{j=1}^K F_j}$, and can thus be defined by function $S_i(x)$:
\begin{equation} \label{eq:Si}
    S_i(x) = \min_{1 \le j\le K}(W_{I}[j,i] \times F_j(x) + (1-W_{I}[j,i]) \times 1).
\end{equation}
\item For the node in the third layer, its shape $T$ is the union of nodes from the second layer, ${{\bigcup_{j=1}^C} S_j}$, and can thus be defined by function $T(x)$:
\begin{equation}\label{eq:Tx}
    T(x) =  \max_{1\le j\le C}(W_{U}[j,1] \times S_j(x) + (1-W_{U}[j,1]) \times 0).
\end{equation}
\end{itemize}

\subsection{Equivalence of CSG-Stump and CSG-Tree}\label{subsec:CSG}
It is easy to verify that a CSG-Stump structure can be converted into a binary CSG-Tree due to the fact that $\cup_{i=1}^n p_i= p_1\cup (p_2\cup (\cdots (p_{n-1}\cup p_n)))$ and $\cap_{i=1}^n p_i = p_1\cap (p_2\cap (\cdots (p_{n-1}\cap p_n)))$ where $p_i$ represent some solid shapes. The reverse is also true. That is, an arbitrary binary CSG-Tree can be represented by a CSG-Stump structure without loss of information, which we prove below. 

In fact, let $P={p_1, p_2, \cdots, p_k}$ be a set of primitive shapes, and  $\otimes = \{\cap, \cup,\setminus\}$ be Boolean operations. We need to prove that a shape defined by a CSG-Tree with primitives $P$ and boolean operations $\otimes$ can be represented by a CSG-Stump structure.

\noindent\textbf{Base Case: }
First we prove that any Boolean operation of 2 primitives $p_1$ and $p_2$ can be represented by a CSG-Stump. This is straightforward since
\begin{equation}
    p_1\cap p_2 = (p_1\cap p_2) \cup \emptyset
\end{equation}
\begin{equation}
    p_1\cup p_2 = (p_1\cap \mathcal{U}) \cup (p_2\cap \mathcal{U})
\end{equation}
\begin{equation}
    p_1\setminus p_2 = (p_1\cap p_2^c) \cup \emptyset
\end{equation}

\noindent\textbf{Inductive Step: }
Next, we assume that any CSG-Tree with less than $n+1$ primitives can be represented by a CSG-Stump. Now consider a shape $\beta$ represented by a CSG-Tree with $n+1$ primitives. The tree is split into two sub-trees at the root node: $\beta = \beta_1 \otimes \beta_2$, where $\beta_1, \beta_2$ represent the shapes defined by the sub-trees and $\otimes$ is the Boolean operation at the root node. Since obviously each of the two sub-trees contains at most $n$ primitives, they can be represented by CSG-Stump. 

Let $\beta_1 = \gamma_1 \cup \cdots \cup \gamma_m$ and 
$\beta_2 = \eta_1 \cup \cdots \cup \eta_h$ where $\gamma_i, \eta_j$ are the intersections of primitives or their complements. We examine the expressions under different Boolean operations.
\begin{itemize}
    \item Union
\begin{equation}
    \beta_1 \cup \beta_2 = \gamma_1 \cup \cdots \cup \gamma_m
    \cup \eta_1 \cup \cdots \cup \eta_h
\end{equation}   
    \item Intersection
\begin{equation}
\begin{array}{lcl}
    \beta_1 \cap \beta_2 &=& (\beta_1\cap \eta_1)\cup \cdots \cup (\beta_1\cap \eta_h)  \\
     &=& (\gamma_1\cap \eta_1)\cup \cdots \cup (\gamma_m\cap \eta_1) \cup  \\
        & & \vdots \\
        & & (\gamma_1\cap \eta_h)\cup \cdots \cup (\gamma_m\cap \eta_h) %
\end{array}
\end{equation}  
    \item Difference
\begin{equation}
\begin{array}{lcl}
    \beta_1 \setminus \beta_2 &=& \beta_1\cap \beta_2^c
    = \beta_1 \cap (\eta_1^c\cap \cdots \cap \eta_h^c)  \\
     &=& (\gamma_1 \cap \eta_1^c\cap \cdots \cap \eta_n^c) \cup  \\
        & & \vdots \\
        & & (\gamma_m \cap \eta_1^c\cap \cdots \cap \eta_h^c)
\end{array}
\end{equation}    
A similar expression can be derived for $\beta_2 \setminus \beta_1$.  
\end{itemize}
All above derivations indicate that $\beta$ can be converted to a CSG-Stump. By mathematical induction, we can conclude that any CSG-tree can be expressed by a CSG-Stump. Hence we theoretically show the equivalence between CSG and CSG-Stump. 

\subsection{Binary Programming Formulation}\label{subsec:MILPForm}
Now let us consider our problem:
the input is a shape given by a point cloud $X=\{x_i\}^N$ consisting of a list of 3D points $x_i$, and we want to reconstruct a CSG-like representation for the shape. With the CSG-Stump, we can come up with a possible solution. We first obtain the target shape occupancy $O_i$  by ~\cite{mescheder2019occupancy} and detect the underlying primitives with a RANSAC-like method~\cite{schnabel2007efficient}. Then reconstructing a CSG-Stump representation is simplified to finding the three connection matrices.
This can be formulated as  a Binary Programming problem. In particular, 
let $O_{k}(i)$ represents the occupancy value of testing point $i$ for primitive $k$ and $T(i)$ represents the estimated occupancy of point $i$. The connection matrices $W_C, W_I$ and $W_U$ for the selection process of CSG-Stump are the solution of the following minimization problem:   
\small
\begin{align} 
\underset{W_{\{C,I,U\}}}{\text{min.}}& \quad   \frac{1}{N}\sum_i^N{\|T(i)-O_i\|}\nonumber\\
\textrm{s.t.}\quad
&T(i)=\max_j\{S_j(i) \times W_{U}[j,1] \}  \nonumber\\
&S_j(i)=\min_k\{F_{k}(i) \times W_{I}[k,j]+(1-W_I[k,j])\}  \nonumber\\
&F_k(i)=(1-O_{k}(i)) W_{C}[1,k] + {O}_{k}(i) (1-W_{C}[1,k])  \nonumber\\
&W_{I}, W_{U}, W_{C}\in\{0,1\},\;\; {O}_{k}(i)\in\{0,1\} 
\vspace{-0.1in}
\end{align} 
\normalsize

\begin{figure}
  \centering
  \includegraphics[width=0.45\textwidth]{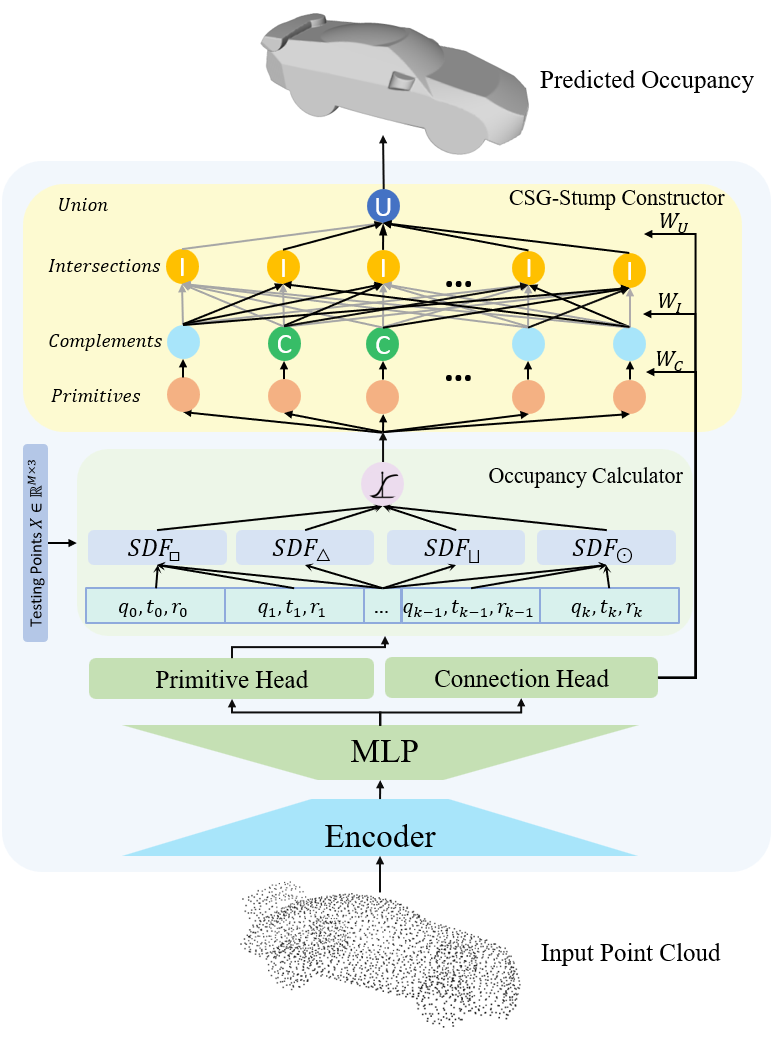}
  \caption{CSG-Stump Net architecture. An input point cloud is fed into an encoder to generate its feature vector. Then the feature vector is decoded by a dual-headed decoder into primitive parameters and CSG-Stump connection weights. Then the Occupancy Calculator computes occupancies on testing points of each predicted primitive. Finally, the CSG-Stump constructor recovers the overall shape occupancy based on predicted connection weights and primitive occupancies.}\label{fig:CSGStumpNet}
  \vspace{-0.2in}
\end{figure}

\section{CSG-Stump Net}\label{sec:CSGStumpNet}
When a relatively large number of primitives are required to represent a shape, Binary Programming typically fails to obtain an optimal solution in polynomial time due to the combinational nature of the problem. We therefore propose a learning-based approach by designing CSG-Stump Net to jointly detect primitives and estimate CSG-Stump connections. As illustrated in Fig.~\ref{fig:CSGStumpNet}, CSG-Stump Net first encodes a point cloud into a latent feature and then decodes it into primitives and connections via the primitive head and the connection head respectively, followed by occupancy calculation and CSG-Stump construction. 

We directly employ an off-the-shelf backbone, i.e. DGCNN~\cite{wang2019dynamic}, as the encoder. Note that our framework is fully compatible with other backbones as well. We discuss the decoder, occupancy calculator, and CSG-Stump constructor in detail as follows.

\subsection{Dual-Headed Decoder}
We first enhance the latent feature with three fully-connected layers ($[512,1024,2048]$), and then use two different heads to further decode the feature into primitive parameters and connection matrices.

\noindent\textbf{Primitive Head.}
Primitive head decodes latent features into a set of $K$ parametric primitives where each parametric primitive is represented by intrinsic and extrinsic parameters. Intrinsic parameters $q$ model the shape of the primitive, such as sphere radius and box dimensions, whereas extrinsic parameters model the global shape transformation composed of a translation vector $t\in\mathbb{R}^3$ and a rotation vector in quaternion form $r\in\mathbb{R}^4$. 
We select four typical types of parametric primitives, i.e., box, sphere, cylinder and cone, as the primitive set, which are standard primitives in CSG representation. For simplicity, we predict equal numbers of $K$ primitives for each type.

\noindent\textbf{CSG-Stump Connection Head.}
CSG-Stump leverages binary matrices to represent Boolean operations among different primitives. We use three dedicated single layer preceptrons to decode the encoded features into the connection matrices $W_C$, $W_I$ and $W_U$. As binary value is not differentiable, we relax this constraint by predicting a soft connection weight in $[0,1]$ using the Sigmoid Function.

\subsection{Differentiable Occupancy Calculator}
To generate primitive's occupancy function in a differential fashion, we first compute the primitive's Signed Distance Field (SDF)~\cite{park2019deepsdf} and then convert it to occupancy~\cite{mescheder2019occupancy} differentially.

Denoting the corresponding operations for the extrinsic parameters of a primitive as translation $T$ and rotation $R$, point $x$ in the world coordinate can be transformed to point $x'$ in a local primitive coordinate as  
$
    x' = T^{-1}(R^{-1}(x)).
$
Afterward, SDF can be calculated according to the mathematical formulation of different primitives. For detailed SDF computation regarding each type of primitives, please refer to the Supplementary Material.

Inspired by \cite{deng2020cvxnet}, SDF is further converted to occupancy by a sigmoid function $\Phi$: \begin{equation} \label{eq:ox}
O(x)=\Phi(-\eta \times SDF(x)),    
\end{equation}  
where the scalar $\eta$ is a hyperparameter indicating the sharpness of the conversion to occupancy.

\subsection{CSG-Stump Constructor} \label{sec:constructor}

Given the predicted primitives occupancy and connection matrices, we can finally calculate the occupancy of the overall shape using the formulations of CSG-Stump described in Sec.~\ref{CSG_Operations}. Note that the complement layer output in~\eqref{eq:Fi} can now be written as:

\begin{align}
    F_i(x) = & W_C[1,i]\times\Phi(\eta\times SDF(x)) + \nonumber\\
        & (1-W_C[1,i])\times\Phi(-\eta\times SDF(x)).
\end{align}

 Though the above CSG-Stump construction process is differentiable and can be directly used in CSG-Stump Net. The gradient can still varnish considering the $\min$ and $\max$ operations only allow gradient back-propagation on the minimal and maximal values. Hence we propose a relaxed version,  $\textit{min*}$ and $\textit{max*}$, using weighted softmax functions:
 \begin{align} \label{eq:smaxmin}
max^*({\boldsymbol{x}}) = \sigma(\psi \cdot {\boldsymbol{x}})\cdot{\boldsymbol{x}}, \\
min^*({\boldsymbol{x}}) = \sigma(-\psi \cdot {\boldsymbol{x}})\cdot{\boldsymbol{x}}, \nonumber
\end{align}
where $\sigma$ is a softmax function and $\psi$ denotes the modulating coefficient.

\subsection{Training and Inference}
\paragraph{Training.}
We train CSG-Stump Net end-to-end in an unsupervised manner. CSG-Stump Net learns to predict a CSG-Stump with primitives and their connections without explicit ground truth. 
Instead, the supervision signal is quantified by the reconstruction loss between the predicted and ground truth occupancy. Specifically, we sample testing points $X\in \mathbb{R}^{N \times 3}$ from the shape bounding box and measure the discrepancy between the ground truth occupancy $O^*$ and the predicted occupancy $\hat{O}$ as follows: 
\begin{equation}
L_{recon} = \mathbb{E}_{x\sim X}||\hat{O}_i-O^*_i||_2^2. 
\end{equation}

In our experiments, we observe that testing points far away from a primitive surface have gradients close to zero, thus stalling the training process. To address this issue, we propose a primitive loss to pull each primitive surface to its closest test point, which prevents the gradient from vanishing. We define this loss term as 
\begin{equation}
L_{primitive} = \frac{1}{K}\sum_k^K \min_{n}SDF_{k}^2(x_n),
\end{equation}
where $SDF_{k}(x_n)$ computes the SDF of test point $x_n$ to primitive $k$.

Finally, the overall objective can be defined as the joint loss of the above two terms: 
\begin{equation}
L_{total} = L_{recon} + \lambda \cdot L_{primitive}, 
\end{equation}
where the balance parameter $\lambda$ is set to $0.001$ empirically. 

\paragraph{Inference.}
During inference, we follow the same procedure as training except that we binarize predicted connection matrices with a threshold $0.5$ to fulfill the binary constraint and generate an interpretable and editable CSG-Stump representation.

\section{Experiments}
In this section, we evaluate CSG-Stump and CSG-Stump Net,  respectively. 
We first demonstrate that our CSG-Stump theoretical framework can achieve optimal solutions by using several toy examples. Then, we evaluate our practical CSG-Stump Net on a large-scale dataset with extensive comparisons and ablation studies. 

\subsection{Evaluate CSG-Stump}
To validate the expressiveness of CSG-Stump, we use the Binary Programming formulation in Sec.~\ref{subsec:MILPForm} to find the optimal solution for CSG-Stump. Particularly, 
to demonstrate the equivalence between CSG-Stump and CSG, we manually create a toy dataset using OpenSCAD~\cite{kintel2014openscad}, which is constructed by a CSG modeling process with different Boolean operations and different primitives. Our dataset consists of six complex shapes, where each shape is constructed with around six primitives with different CSG-Tree. 
The details of the dataset can be found in the supplementary. 

We optimize the problem by randomly sampling $N=1000$ testing points inside the shape bounding box. 
To get rid of the influence of primitive detection and target occupancy estimation, the input primitive occupancy $\hat{O}_{ik}$ for point $i$ and primitive $k$ is directly calculated on the ground truth primitives, while the target occupancy $O_i$ is computed from the target shape. 
We optimize the binary CSG-Stump weight variables, i.e., $W_C$, $W_I$, and $W_U$, by using the off-the-shelf Gurobi solver~\cite{optimization2014inc}.

\begin{figure}[t]
  \centering
  \includegraphics[width=0.45\textwidth]{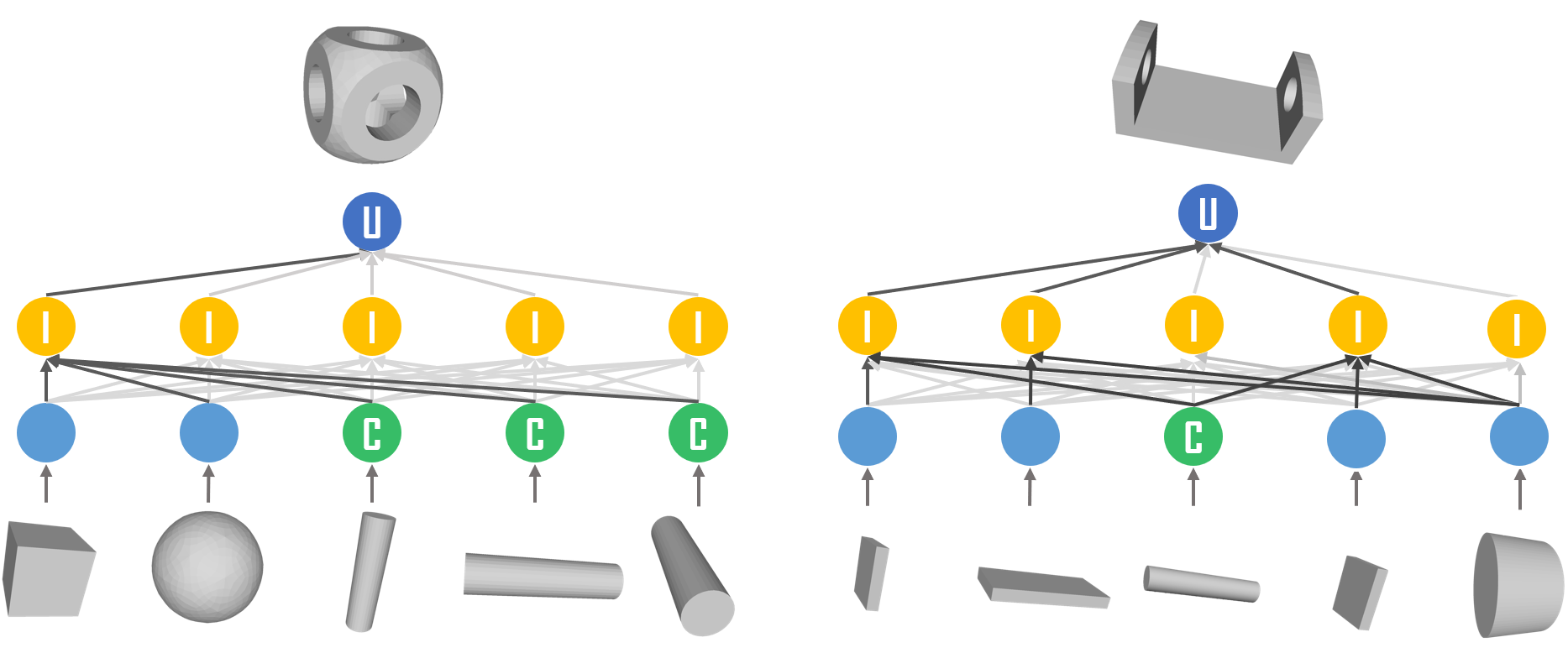}
  \caption{Examples of our estimated CSG-Stump connections, where nodes ``I'', ``U'' and ``C''  represent intersection, union and shape complement.}\label{fg:CSGS-connect}
  \vspace{-0.1in}
\end{figure}

In our experiments, the solver can find the optimal solution and converge to zero objective loss within a minute for the toy dataset. Two CSG-Stump parsed examples are shown in Fig.~\ref{fg:CSGS-connect}.
However, the solver fails to find an optimal solution within a reasonable time limit when we test on some complex shapes from ShapeNet Dataset~\cite{chang2015shapenet}. This is likely because of the combinatorial complexity of the optimization problem and noisy input shapes. Therefore, we propose a deep learning based CSG-Stump Net solution.

\begin{figure*}[h]
  \centering
  \includegraphics[width=0.98\textwidth]{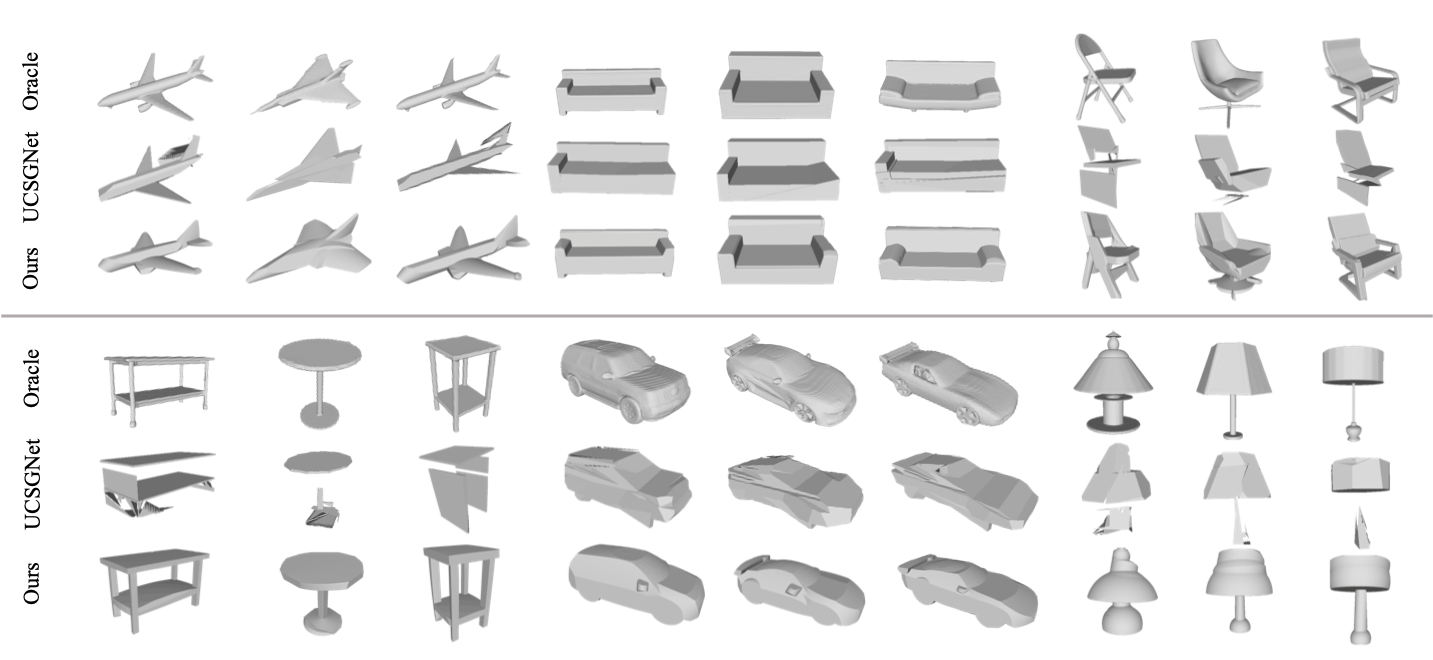}
  \caption{Comparisons of the reconstruction results of CSG-Stump Net and UCSG-Net~\cite{kania2020ucsg}.} \label{fg:recon}
  \vspace{-0.1in}
\end{figure*}

\subsection{Evaluate CSG-Stump Net}
\noindent\textbf{Dataset.} 
We evaluate CSG-Stump Net on ShapeNet dataset~\cite{chang2015shapenet} with the standard splits. 
We randomly sample $2048$ points on a shape surface as an input point cloud and generate $N=2048$ points in the shape bounding box as testing points. The target occupancy for testing points are obtained following \cite{mescheder2019occupancy}.
In our experiments, we found that randomly sampled testing points tend to miss reconstructing thin structures.
Therefore, we use a balanced sampling strategy ($1:1$ for inside and outside points) during training.

\noindent
\textbf{Implementation Details.}
CSG-Stump Net is implemented in Pytorch \cite{paszke2019pytorch} and is optimized with the Adam solver with a learning rate of $10^{-4}$. We distributively train the network on $16$ nVIDIA V100 32GB GPUs with a batch size of $32$. It took about one week to converge on all $13$ classes. 

In our experiments, we set $K=256$ and $C=256$ for the intersection layer. We demonstrate their impacts on results in the ablation studies.
As most shapes in ShapeNet dataset can be constructed with just intersection and union, we directly set complement weight $W_C$ to zeros. We also try to learn a dynamic complement weight $W_C$, which results in a slightly better performance with more learning space and higher computational burden. 
For hyper-parameters in~\eqref{eq:ox} and~\eqref{eq:smaxmin}, we set $\sigma=75$ and $\psi=20$, empirically. 

\noindent\textbf{Comparisons.}
We compare our method with both CSG-like methods, i.e. UCSG-Net~\cite{kania2020ucsg}, and primitive decomposition methods, including VP~\cite{tulsiani2017learning} and SQ~\cite{paschalidou2019superquadrics}. 
We evaluate results on the $L_2$ Chamfer Distance between $2048$ sampled points on a reconstructed shape and those on the corresponding ground truth following UCSG-Net. 
The quantitative results are reported in Table~\ref{tab:recon}. 
Note that the results of VP, SQ and UCSGNet are those reported in UCSG-Net. 
We can see that CSG-Stump Net outperforms both kinds of methods and improve previous SOTA results by over $9\%$. 
Fig.~\ref{fg:recon} shows the qualitative comparison with the CSG-like counterpart, i.e. UCSG-Net. 
We can see that our method achieves much better geometry approximation and structure decomposition in comparison to oracle shapes.  
\begin{table}[h]
  \centering
  \caption{3D Reconstruction quantitative results measured by $L_2$ Chamfer Distance (CD) on ShapeNet Dataset. Our CSG-Stump Net outperforms the baselines by convincing margins. The CD values are  multiplied by 1000 for easy reading.}
  \label{tab:recon}
  \begin{tabular}{c c c c c}
\hline
      & VP~\cite{tulsiani2017learning} & SQ~\cite{paschalidou2019superquadrics} & UCSGNet~\cite{kania2020ucsg} & Ours\\
  \hline
    CD & 2.259 & 1.656 & 2.085 & \textbf{1.505} \\
  \hline
  \end{tabular}
  \vspace{-0.1in}
\end{table}

\begin{figure}[h]
  \centering
  \includegraphics[width=0.45\textwidth]{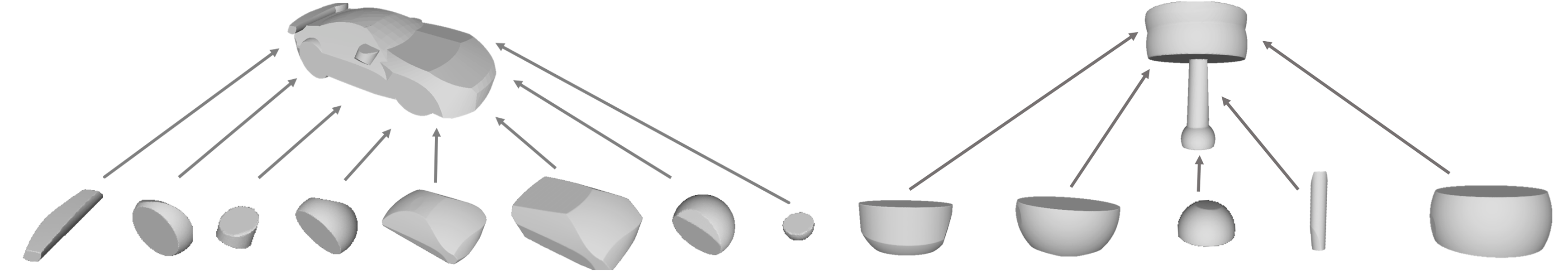}
  \caption{Two examples show how a car and a lamp can be constructed with a small number of parts of intersection nodes. The decomposition is compact, interpretable, and consistent with the semantic parts of cars and lamps. Note that we only show the visible parts.} \label{fg:compact}
  \vspace{-0.1in}
\end{figure}

\noindent\textbf{Performance on Compactness.} Fig.~\ref{fg:compact} shows the generated CSG-stump structures of the car and lamp examples. We can see that only a small subset of intersection nodes is used to construct the final shape, which suggests that the obtained structure is compact.  Interestingly, the automatically learned intersection nodes are consistent with the semantic part decomposition of the car and lamp, which may be useful for other tasks such as part segmentation. 

\begin{figure}[h]
  \centering
  \includegraphics[width=0.45\textwidth]{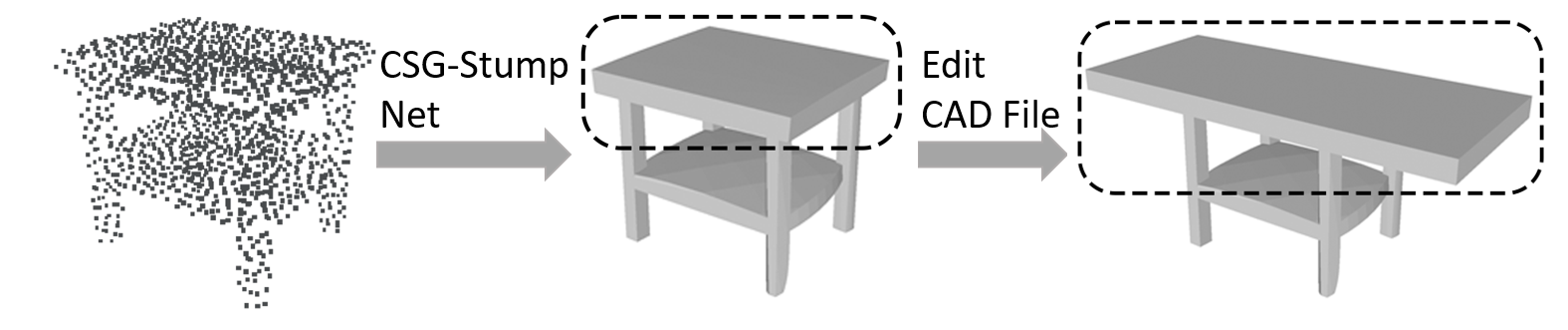}
  \caption{A editing workflow. A CAD-compatible shape is firstly recovered from a point cloud by CSG-Stump Net, then users can edit it in a CAD software for novel designs.} \label{fg:edit}
  \vspace{-0.1in}
\end{figure}

\noindent\textbf{Performance on Editability.}
As CSG-Stump is equivalent to CSG, we are allowed to edit primitives and CSG-Stump connections for further designs.   
Specifically, we implemented a simple adaptor to convert our outputs to the input format of OpenSCAD files, where OpenSCAD is an open-sourced CAD software. 
By leveraging OpenSCAD's editing user interface and CSG-Stump Net, a user can achieve a design aim directly based on a point cloud (see Fig.~\ref{fg:edit}).

\noindent \textbf{Ablation on the Number of Primitives.}
The max number of available primitives is an important factor of CSG-Stump. Intuitively, more available primitives lead to a better approximation. 
However, too many primitives can make results complex and not editable as well as increase the network complexity and inference computation. 
Table~\ref{tab:num-primitive} shows CD results under different numbers of primitives. We can see that allowing more primitives improves the performance.

\noindent\textbf{Ablation on the Number of Intersection Nodes.}
Apart from the number of available primitives, the number of intersection nodes also affects the overall quality of results. Table~\ref{tab:num-intersec} shows the CD results under different numbers of intersection nodes. In general, more intersection nodes lead to better results but at the cost of reducing compactness and increasing network complexity. 

\begin{table}[h]
  \centering
  \caption{Chamfer Distances of Airplane class under different numbers of available primitives.} 
  \label{tab:num-primitive}
  \begin{tabular}{c c c c c}
  \hline
   \# Primitives   & 256 & 128 & 64 & 32 \\
  \hline      
   CD & \textbf{1.22} & 1.28 & 1.40 & 1.44 \\
  \hline  
  \end{tabular}
  \vspace{-0.1in}
\end{table}

\begin{table}[h]
  \centering
  \caption{Chamfer Distances of Airplane class under different numbers of intersection nodes.}
  \label{tab:num-intersec}
  \begin{tabular}{c c c c c}
  \hline
   \# Intersection Nodes  & 256 & 128 & 64 & 32 \\
  \hline      
   CD & \textbf{1.22} & 1.37 & 2.28 & 2.26 \\
  \hline  
  \end{tabular}
  \vspace{-0.1in}
\end{table}

\noindent\textbf{Ablation on the Type of Primitives.}
Experience in CAD modeling has shown that more kinds of primitives can increase the modeling ability of CSG while increasing complexity in modeling software. 
We study how different available types of primitives affect overall results quantitatively in Table 4 and qualitatively in Fig.~7. 
We can see that CSG-Stump Net can well approximate a shape with different kinds of primitives.

\begin{figure}[h]
  \centering
  \includegraphics[width=0.5\textwidth]{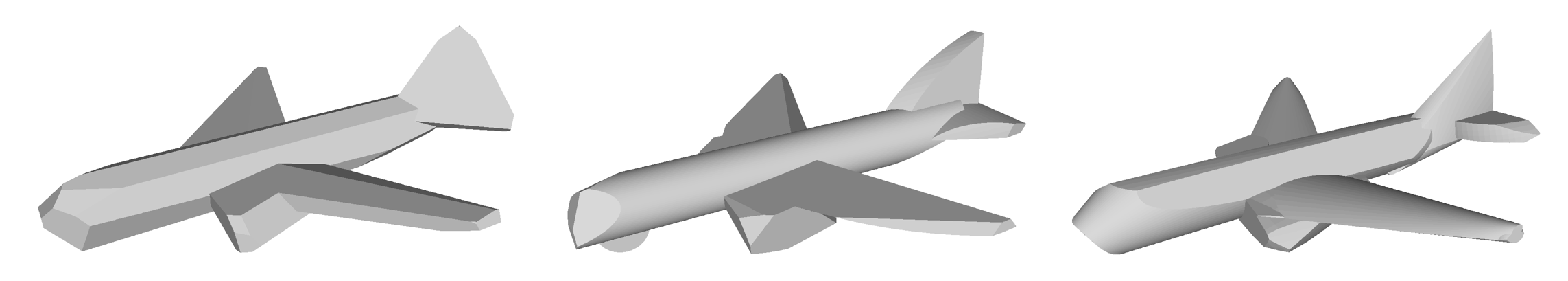}
  \caption{\textbf{Different types of Primitives:} From left to right are the results only using box, using box and cylinder (see fuselage), and using all types (e.g. cone for airplane nose), respectively.}
\end{figure}

\begin{table}[h]
  \centering
  \caption{Chamfer Distance (CD) results of Airplane class with different primitive types.}
  \label{tab:1}
  \begin{tabular}{c c c c c}
  \hline 
      & Box & Box Cylinder & All Types \\
  \hline      
  CD & 1.27 & 1.24 & 1.22 \\
  \hline  
  \end{tabular}
  \vspace{-0.1in}
\end{table}


\section{Conclusion}
We have presented CSG-Stump, a three-level CSG-like representation for 3D shapes. While it inherits the compact, interpretable and editable nature of CSG-Tree, it is learning-friendly and has high representation capability. Based on CSG-Stump, we design CSG-Stump Net, which can be trained end-to-end in an unsupervised manner. We demonstrate through extensive experiments that CSG-Stump out-performs existing methods by a significant margin.

\clearpage

{\small
\bibliographystyle{ieee_fullname}
\bibliography{CSGStump.bib}

\begin{thebibliography}{10}\itemsep=-1pt

\bibitem{bronstein2017geometric}
Michael~M Bronstein, Joan Bruna, Yann LeCun, Arthur Szlam, and Pierre
  Vandergheynst.
\newblock Geometric deep learning: going beyond euclidean data.
\newblock {\em IEEE Signal Processing Magazine}, 34(4):18--42, 2017.

\bibitem{chang2015shapenet}
Angel~X Chang, Thomas Funkhouser, Leonidas Guibas, Pat Hanrahan, Qixing Huang,
  Zimo Li, Silvio Savarese, Manolis Savva, Shuran Song, Hao Su, et~al.
\newblock Shapenet: An information-rich 3d model repository.
\newblock {\em arXiv preprint arXiv:1512.03012}, 2015.

\bibitem{chen2020bsp}
Zhiqin Chen, Andrea Tagliasacchi, and Hao Zhang.
\newblock Bsp-net: Generating compact meshes via binary space partitioning.
\newblock In {\em Proceedings of the IEEE/CVF Conference on Computer Vision and
  Pattern Recognition}, pages 45--54, 2020.

\bibitem{chen2019bae}
Zhiqin Chen, Kangxue Yin, Matthew Fisher, Siddhartha Chaudhuri, and Hao Zhang.
\newblock Bae-net: Branched autoencoder for shape co-segmentation.
\newblock In {\em Proceedings of the IEEE/CVF International Conference on
  Computer Vision}, pages 8490--8499, 2019.

\bibitem{chen2019learning}
Zhiqin Chen and Hao Zhang.
\newblock Learning implicit fields for generative shape modeling.
\newblock In {\em Proceedings of the IEEE/CVF Conference on Computer Vision and
  Pattern Recognition}, pages 5939--5948, 2019.

\bibitem{choy20163d}
Christopher~B Choy, Danfei Xu, JunYoung Gwak, Kevin Chen, and Silvio Savarese.
\newblock 3d-r2n2: A unified approach for single and multi-view 3d object
  reconstruction.
\newblock In {\em European conference on computer vision}, pages 628--644.
  Springer, 2016.

\bibitem{deng2020cvxnet}
Boyang Deng, Kyle Genova, Soroosh Yazdani, Sofien Bouaziz, Geoffrey Hinton, and
  Andrea Tagliasacchi.
\newblock Cvxnet: Learnable convex decomposition.
\newblock In {\em Proceedings of the IEEE/CVF Conference on Computer Vision and
  Pattern Recognition}, pages 31--44, 2020.

\bibitem{goldfeather1989near}
Jack Goldfeather, Steven Monar, Greg Turk, and Henry Fuchs.
\newblock Near real-time csg rendering using tree normalization and geometric
  pruning.
\newblock {\em IEEE Computer Graphics and Applications}, 9(3):20--28, 1989.

\bibitem{guo20153d}
Kan Guo, Dongqing Zou, and Xiaowu Chen.
\newblock 3d mesh labeling via deep convolutional neural networks.
\newblock {\em ACM Transactions on Graphics (TOG)}, 35(1):1--12, 2015.

\bibitem{hao2020dualsdf}
Zekun Hao, Hadar Averbuch-Elor, Noah Snavely, and Serge Belongie.
\newblock Dualsdf: Semantic shape manipulation using a two-level
  representation.
\newblock In {\em Proceedings of the IEEE/CVF Conference on Computer Vision and
  Pattern Recognition}, pages 7631--7641, 2020.

\bibitem{kania2020ucsg}
Kacper Kania, Maciej Zi{\k{e}}ba, and Tomasz Kajdanowicz.
\newblock Ucsg-net--unsupervised discovering of constructive solid geometry
  tree.
\newblock {\em arXiv preprint arXiv:2006.09102}, 2020.

\bibitem{kintel2014openscad}
Marius Kintel and Clifford Wolf.
\newblock Openscad.
\newblock {\em GNU General Public License, p GNU General Public License}, 2014.

\bibitem{laidlaw1986constructive}
David~H Laidlaw, W~Benjamin Trumbore, and John~F Hughes.
\newblock Constructive solid geometry for polyhedral objects.
\newblock In {\em Proceedings of the 13th annual conference on Computer
  graphics and interactive techniques}, pages 161--170, 1986.

\bibitem{li2019supervised}
Lingxiao Li, Minhyuk Sung, Anastasia Dubrovina, Li Yi, and Leonidas~J Guibas.
\newblock Supervised fitting of geometric primitives to 3d point clouds.
\newblock In {\em Proceedings of the IEEE/CVF Conference on Computer Vision and
  Pattern Recognition}, pages 2652--2660, 2019.

\bibitem{LU20171}
Yang Lu.
\newblock Industry 4.0: A survey on technologies, applications and open
  research issues.
\newblock {\em Journal of Industrial Information Integration}, 6:1--10, 2017.

\bibitem{mescheder2019occupancy}
Lars Mescheder, Michael Oechsle, Michael Niemeyer, Sebastian Nowozin, and
  Andreas Geiger.
\newblock Occupancy networks: Learning 3d reconstruction in function space.
\newblock In {\em Proceedings of the IEEE/CVF Conference on Computer Vision and
  Pattern Recognition}, pages 4460--4470, 2019.

\bibitem{mildenhall2020nerf}
Ben Mildenhall, Pratul~P Srinivasan, Matthew Tancik, Jonathan~T Barron, Ravi
  Ramamoorthi, and Ren Ng.
\newblock Nerf: Representing scenes as neural radiance fields for view
  synthesis.
\newblock In {\em European Conference on Computer Vision}, pages 405--421.
  Springer, 2020.

\bibitem{optimization2014inc}
Gurobi Optimization.
\newblock Inc.,“gurobi optimizer reference manual,” 2015, 2014.

\bibitem{park2019deepsdf}
Jeong~Joon Park, Peter Florence, Julian Straub, Richard Newcombe, and Steven
  Lovegrove.
\newblock Deepsdf: Learning continuous signed distance functions for shape
  representation.
\newblock In {\em Proceedings of the IEEE/CVF Conference on Computer Vision and
  Pattern Recognition}, pages 165--174, 2019.

\bibitem{paschalidou2019superquadrics}
Despoina Paschalidou, Ali~Osman Ulusoy, and Andreas Geiger.
\newblock Superquadrics revisited: Learning 3d shape parsing beyond cuboids.
\newblock In {\em Proceedings of the IEEE/CVF Conference on Computer Vision and
  Pattern Recognition}, pages 10344--10353, 2019.

\bibitem{paszke2019pytorch}
Adam Paszke, Sam Gross, Francisco Massa, Adam Lerer, James Bradbury, Gregory
  Chanan, Trevor Killeen, Zeming Lin, Natalia Gimelshein, Luca Antiga, et~al.
\newblock Pytorch: An imperative style, high-performance deep learning library.
\newblock {\em arXiv preprint arXiv:1912.01703}, 2019.

\bibitem{pontes2018image2mesh}
Jhony~K Pontes, Chen Kong, Sridha Sridharan, Simon Lucey, Anders Eriksson, and
  Clinton Fookes.
\newblock Image2mesh: A learning framework for single image 3d reconstruction.
\newblock In {\em Asian Conference on Computer Vision}, pages 365--381.
  Springer, 2018.

\bibitem{qi2017pointnet}
Charles~R Qi, Hao Su, Kaichun Mo, and Leonidas~J Guibas.
\newblock Pointnet: Deep learning on point sets for 3d classification and
  segmentation.
\newblock In {\em Proceedings of the IEEE conference on computer vision and
  pattern recognition}, pages 652--660, 2017.

\bibitem{qi2017pointnet++}
Charles~R Qi, Li Yi, Hao Su, and Leonidas~J Guibas.
\newblock Pointnet++: Deep hierarchical feature learning on point sets in a
  metric space.
\newblock {\em arXiv preprint arXiv:1706.02413}, 2017.

\bibitem{riegler2017octnet}
Gernot Riegler, Ali Osman~Ulusoy, and Andreas Geiger.
\newblock Octnet: Learning deep 3d representations at high resolutions.
\newblock In {\em Proceedings of the IEEE conference on computer vision and
  pattern recognition}, pages 3577--3586, 2017.

\bibitem{saito2019pifu}
Shunsuke Saito, Zeng Huang, Ryota Natsume, Shigeo Morishima, Angjoo Kanazawa,
  and Hao Li.
\newblock Pifu: Pixel-aligned implicit function for high-resolution clothed
  human digitization.
\newblock In {\em Proceedings of the IEEE/CVF International Conference on
  Computer Vision}, pages 2304--2314, 2019.

\bibitem{schnabel2007efficient}
Ruwen Schnabel, Roland Wahl, and Reinhard Klein.
\newblock Efficient ransac for point-cloud shape detection.
\newblock In {\em Computer graphics forum}, volume~26, pages 214--226. Wiley
  Online Library, 2007.

\bibitem{sharma2018csgnet}
Gopal Sharma, Rishabh Goyal, Difan Liu, Evangelos Kalogerakis, and Subhransu
  Maji.
\newblock Csgnet: Neural shape parser for constructive solid geometry.
\newblock In {\em Proceedings of the IEEE Conference on Computer Vision and
  Pattern Recognition}, pages 5515--5523, 2018.

\bibitem{solidworks2005solidworks}
Dassault~Syst{\`e}mes SolidWorks.
\newblock Solidworks{\textregistered}.
\newblock {\em Version Solidworks}, 2005.

\bibitem{thomas2019kpconv}
Hugues Thomas, Charles~R Qi, Jean-Emmanuel Deschaud, Beatriz Marcotegui,
  Fran{\c{c}}ois Goulette, and Leonidas~J Guibas.
\newblock Kpconv: Flexible and deformable convolution for point clouds.
\newblock In {\em Proceedings of the IEEE/CVF International Conference on
  Computer Vision}, pages 6411--6420, 2019.

\bibitem{tulsiani2017learning}
Shubham Tulsiani, Hao Su, Leonidas~J Guibas, Alexei~A Efros, and Jitendra
  Malik.
\newblock Learning shape abstractions by assembling volumetric primitives.
\newblock In {\em Proceedings of the IEEE Conference on Computer Vision and
  Pattern Recognition}, pages 2635--2643, 2017.

\bibitem{wang2018pixel2mesh}
Nanyang Wang, Yinda Zhang, Zhuwen Li, Yanwei Fu, Wei Liu, and Yu-Gang Jiang.
\newblock Pixel2mesh: Generating 3d mesh models from single rgb images.
\newblock In {\em Proceedings of the European Conference on Computer Vision
  (ECCV)}, pages 52--67, 2018.

\bibitem{wang2017cnn}
Peng-Shuai Wang, Yang Liu, Yu-Xiao Guo, Chun-Yu Sun, and Xin Tong.
\newblock O-cnn: Octree-based convolutional neural networks for 3d shape
  analysis.
\newblock {\em ACM Transactions on Graphics (TOG)}, 36(4):1--11, 2017.

\bibitem{wang2018adaptive}
Peng-Shuai Wang, Chun-Yu Sun, Yang Liu, and Xin Tong.
\newblock Adaptive o-cnn: a patch-based deep representation of 3d shapes.
\newblock {\em ACM Transactions on Graphics (TOG)}, 37(6):1--11, 2018.

\bibitem{wang2019dynamic}
Yue Wang, Yongbin Sun, Ziwei Liu, Sanjay~E Sarma, Michael~M Bronstein, and
  Justin~M Solomon.
\newblock Dynamic graph cnn for learning on point clouds.
\newblock {\em Acm Transactions On Graphics (tog)}, 38(5):1--12, 2019.

\bibitem{wen2019pixel2mesh++}
Chao Wen, Yinda Zhang, Zhuwen Li, and Yanwei Fu.
\newblock Pixel2mesh++: Multi-view 3d mesh generation via deformation.
\newblock In {\em Proceedings of the IEEE/CVF International Conference on
  Computer Vision}, pages 1042--1051, 2019.

\bibitem{wu20153d}
Zhirong Wu, Shuran Song, Aditya Khosla, Fisher Yu, Linguang Zhang, Xiaoou Tang,
  and Jianxiong Xiao.
\newblock 3d shapenets: A deep representation for volumetric shapes.
\newblock In {\em Proceedings of the IEEE conference on computer vision and
  pattern recognition}, pages 1912--1920, 2015.

\end{thebibliography}
}

\clearpage
\onecolumn
\begin{center}
  {\Large \bf {CSG-Stump: A Learning Friendly CSG-Like Representation \\ for Interpretable Shape Parsing} \\
  \vspace{4mm}
-- Supplementary Materials --
}
 \end{center}
\setcounter{section}{0}
\setcounter{figure}{0}
\def\thesection{\Alph{section}}
\def\thefigure{\Alph{figure}}
\section{Signed-Distance-Field Calculation}
To compute the signed distance of a point $x$ with respect to a primitive $p$, we first transform the point $x$ from the world coordinate system into $x'$ in primitive's local coordinate system as discussed in Sec.4.2 of the paper.

\paragraph{Box}
We define the origin of a box's local coordinate system as the box center. Its shape is defined by a 3 dimensional positive vector $Q_{Box}=\langle Q_{Box}[0],Q_{Box}[1],Q_{Box}[2] \rangle$ indicating the box's width, height and depth. Since our defined boxes are symmetric about the coordinate system's $x$\nobreakdash--$y$ $x$\nobreakdash--$z$ and $y$\nobreakdash--$z$~planes, we can transform the point $x'$ to the first octant by $|x'|=\langle |x'_x|, |x'_y|, |x'_z| \rangle$ without changing its signed distance. We also define utility functions $f_{max}(\vec{x}, a)= \langle \max(x_0, a), \max(x_1, a), \max(x_2, a) \rangle$ and $g_{max}(\vec{x}, a)= \max(x_0, x_1, x_2, a)$ to ease our discussion.

\noindent Thus we can compute the Signed-Distance Field of a box $SDF_\square$ as follows:

\begin{align}
 SDF_{\square}(x')= & ||f_{max}(|x'|- 0.5 \cdot Q_{box},0)||_2 +\min(g_{max}(|x'|-0.5 \cdot Q_{box}), 0)
\end{align}

\noindent Note that the first term is in charge of computing SDF when a point is outside of the box, and the second term for inside points. 

\paragraph{Sphere}
Similarly, we define the origin of a sphere's local coordinate system as the sphere center. The shape of a sphere is defined by a positive scalar $Q_{sphere}$ indicating its radius. Thus the Signed-Distance-Field of a sphere $SDF_{\bigcirc}$ can be defined as follows:

\begin{equation}
    SDF_{\bigcirc}(x')=||x'||_2 - Q_{sphere}
\end{equation}

\paragraph{Cylinder}
We define the z-axis of a cylinder's local coordinate system as the cylinder's axis. The shape of a infinitely long cylinder is defined by a positive scalar $Q_{cylinder}$ indicating its radius. Thus the Signed-Distance-Field of a cylinder $SDF_{\bigsqcup}$ can be defined as follows:

\begin{equation}
    SDF_{\bigsqcup}(x')=||x'_{x,y}||_2 - Q_{cylinder}
\end{equation}

\paragraph{Cone}
We define the origin of a cone's local coordinate system as a cone's apex, and the cone is extending downwards along the z-axes. The shape of a cone extends infinitely far is defined by a positive scalar $Q_{cone}$ indicating its opening angle. Thus the Signed-Distance-Field of a cone $SDF_{\triangle}$ can be defined as follows:

\begin{equation}
    SDF_{\triangle}(x')=
    \begin{cases}
    ||x'||_2 & \text{if } x'_{z} \geq 0\\
    \frac{(||x'_{x,y}||_2 - x'_{z} \cdot \tan(Q_{cone})}{\sqrt{1+\tan^2(Q_{cone})}} & \text{otherwise}
    \end{cases}
\end{equation}

\section{Toy Dataset}
Apart from the two shapes shown in Fig.3 of the paper, the rest of the toy dataset is shown in Fig.A, which consists of multiple complex shapes defined using a multilevel CSG-Parse-Tree with different boolean operations and different types of primitives.  

\section{A Complete CSG-Stump Example (16 $\times$ 16)}
We show a raw output of CSG-Stump Net in Fig.B. We plot the complete CSG-Stump with estimated primitives and connection matrices for an airplane shape trained with 16 primitives and 16 intersection nodes. 

\section{Visual Results}
We show more generated results in detail in Fig.C. Note that all meshes are parametric shapes exported using openSCAD. We also include a video showing the generated shapes in 360 degree views.

\begin{figure*}[h]
  \centering
  \includegraphics[width=0.9\textwidth]{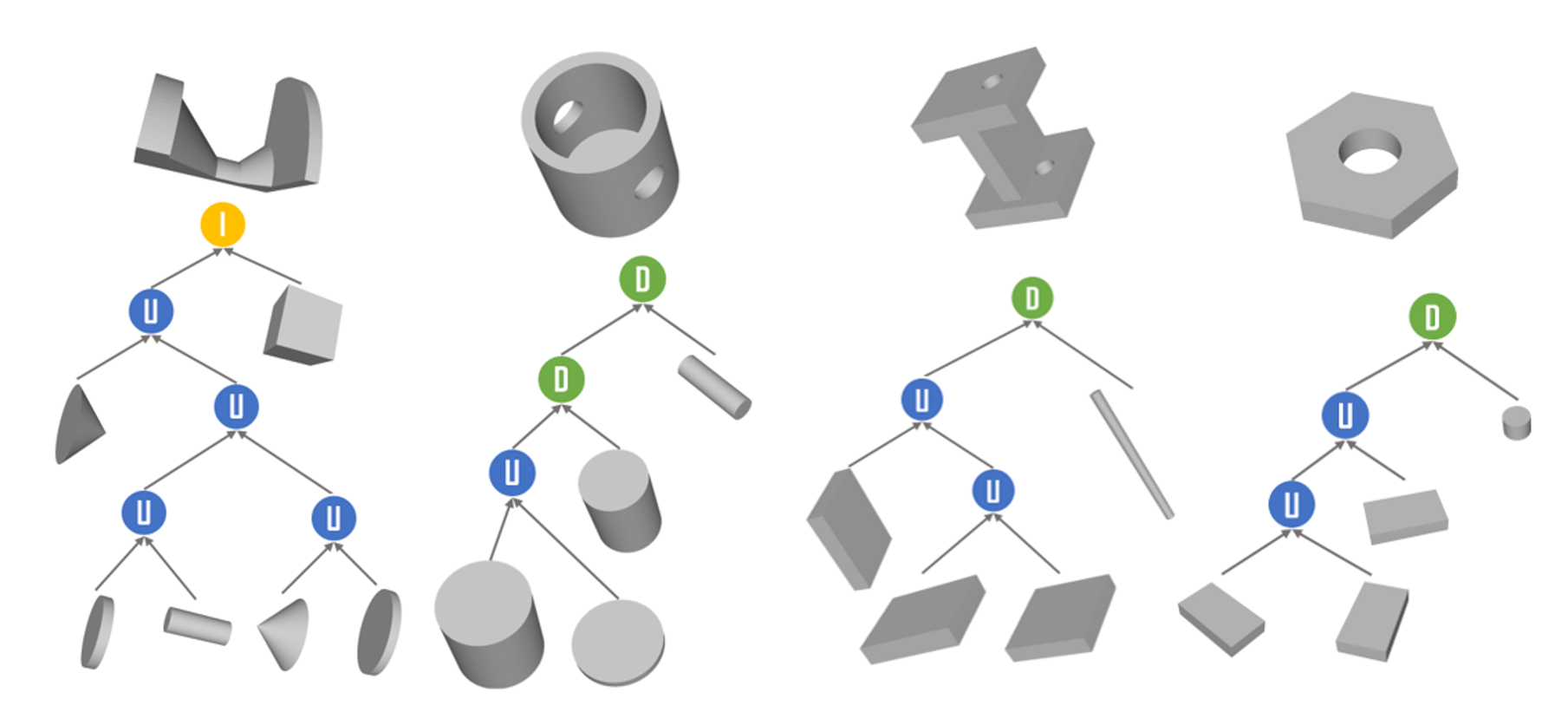}
  \caption{The toy dataset for CSG-Stump Optimization.}
\end{figure*}\label{fig:toy_datasets}

\begin{figure*}[h]
  \centering
  \includegraphics[width=0.9\textwidth]{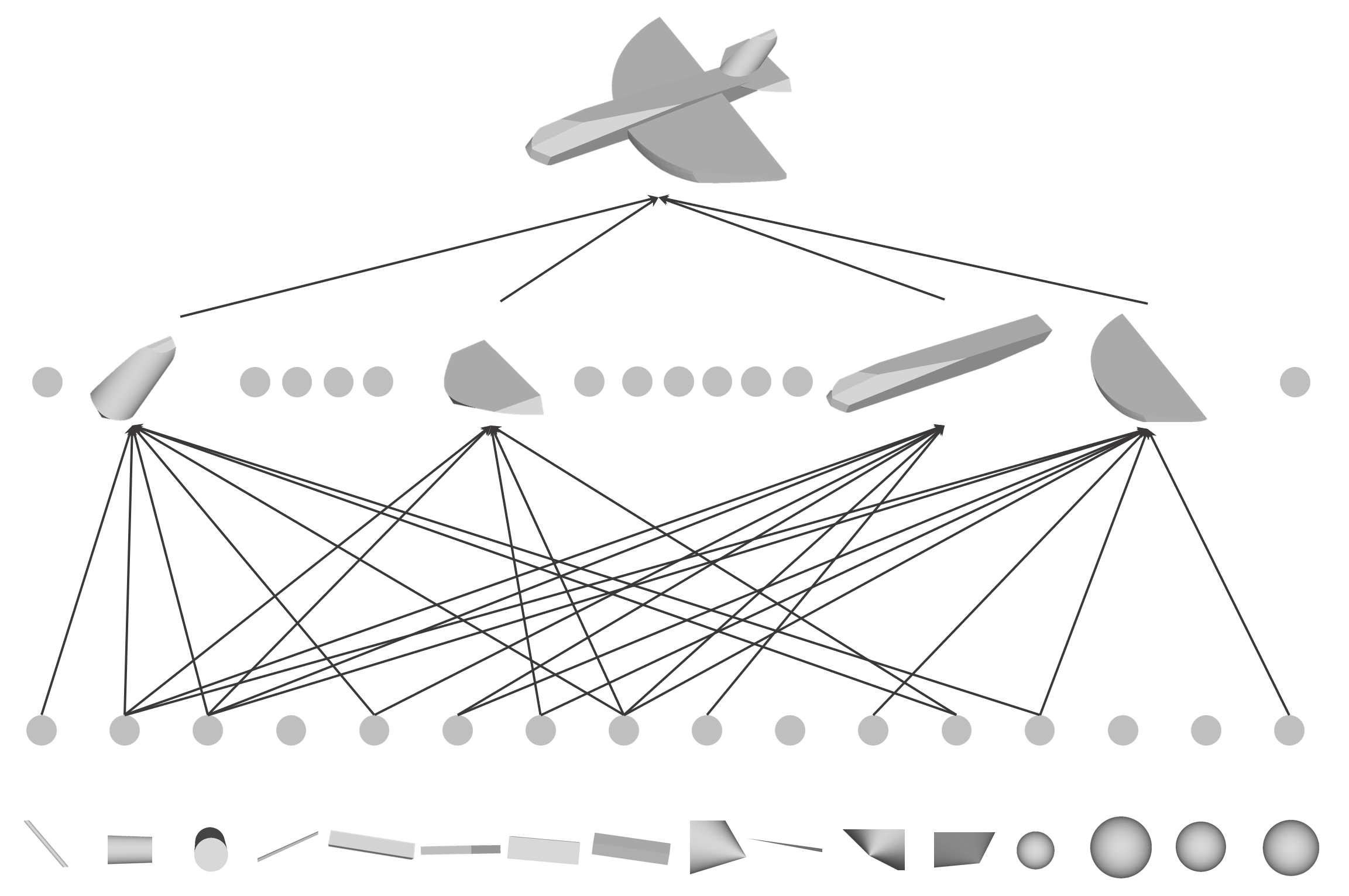}
  \caption{Our predicted CSG-Stump structure with 16 primitives. Note that this CSG-Stump is directly outputted from our model without any modification and simplification.}
\end{figure*}\label{fig:CSG-Stump_complete}

\begin{figure*}[h]
  \centering
  \includegraphics[width=\textwidth]{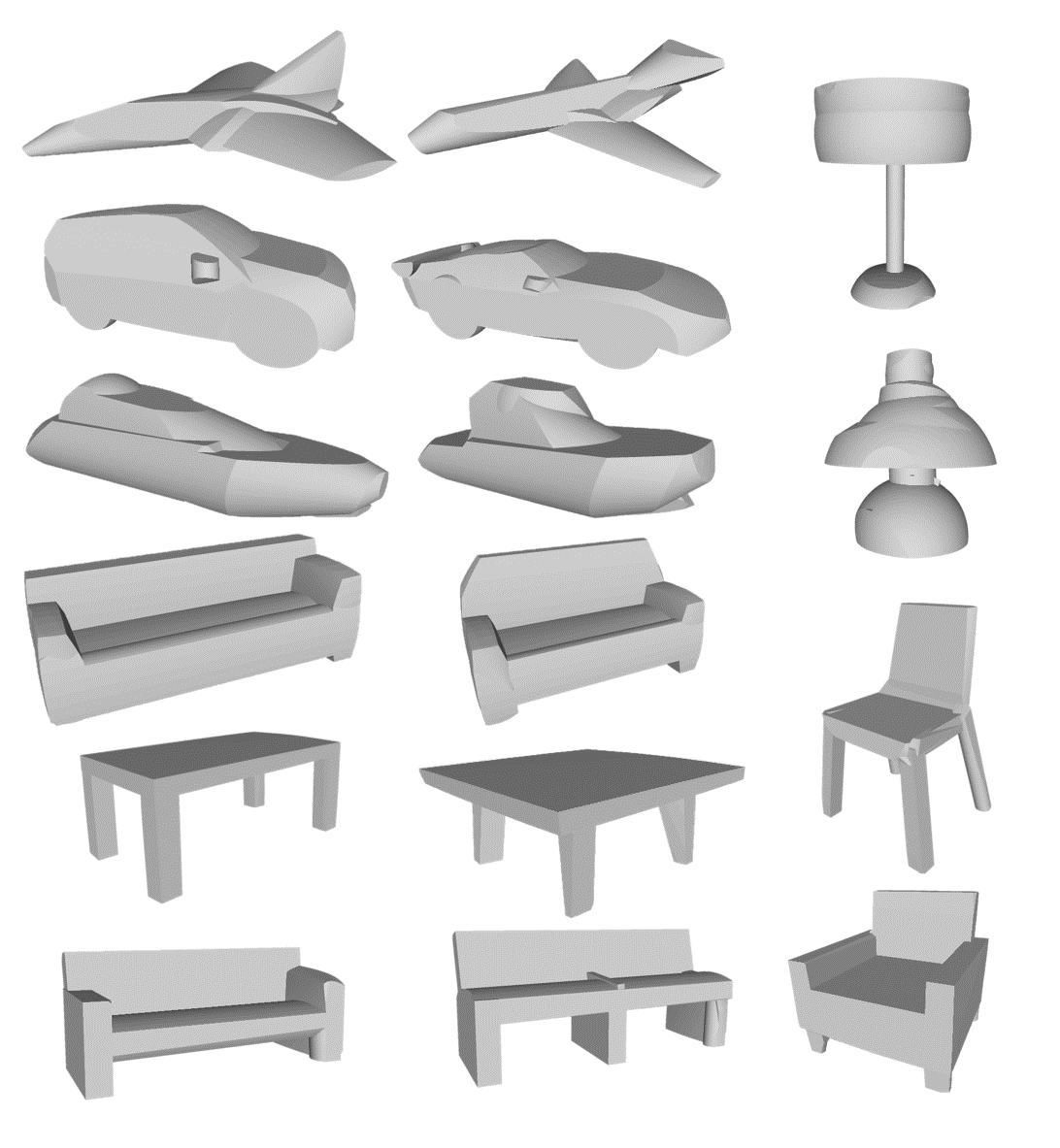}
  \caption{Our generated shapes rendered by CAD Software.}\label{fig:generation}
\end{figure*}

\end{document}